\documentclass[runningheads]{llncs}

\usepackage{eccv}

\usepackage{eccvabbrv}

\usepackage{graphicx}
\usepackage{booktabs}
\usepackage{multirow}
\usepackage{tabularx} 

\usepackage[accsupp]{axessibility}  

\usepackage{hyperref}

\usepackage{orcidlink}

\begin{document}

\title{Temporal-consistent CAMs for Weakly Supervised Video Segmentation in Waste Sorting} 

\titlerunning{Temporal-consistent CAMs}

\author{Andrea Marelli
\orcidlink{0009-0009-4418-110X}
\and
Luca Magri
\orcidlink{0000-0002-0598-8279} \and
Federica Arrigoni
\orcidlink{0000-0003-0331-4032} \and
Giacomo Boracchi
\orcidlink{0000-0002-1650-3054}}

\authorrunning{A.~Marelli et al.}

\institute{DEIB - Politecnico di Milano, Italy}

\maketitle

\begin{abstract}
In industrial settings, weakly supervised (WS) methods are usually preferred over their fully supervised (FS) counterparts as they do not require costly manual annotations. Unfortunately, the segmentation masks obtained in the WS regime are typically poor in terms of accuracy. In this work, we present a WS method capable of producing accurate masks for semantic segmentation in case of video streams. More specifically, we build saliency maps that exploit the temporal coherence between consecutive frames in a video, promoting consistency when objects appear in different frames. 
We apply our method in a waste-sorting scenario, where we perform weakly supervised video segmentation (WSVS) by training an auxiliary classifier that distinguishes between videos recorded before and after a human operator, who manually removes specific wastes from a conveyor belt. The saliency maps of this classifier identify materials to be removed, and we modify the classifier training to minimize differences between the saliency map of a central frame and those in adjacent frames, after having compensated object displacement. Experiments on a real-world dataset demonstrate the benefits of integrating temporal coherence directly during the training phase of the classifier. Code and dataset are available upon request.
  
\end{abstract}

\section{Introduction}
\label{sec:intro}

With the escalation of global waste production, it has become critical to improve modern waste management systems. In particular, waste sorting involves the separation of specific types of recyclable waste, which usually consists in manually removing objects of different materials from a conveyor belt, where only a specific material must remain. Machine vision systems and deep learning have emerged as promising solutions to automatize these processes, aiming to enhance the efficiency and accuracy of waste management to reduce human error and to lower operational costs\cite{s23073646, GUNDUPALLI201756}. These advancements significantly contribute to more sustainable and environmentally friendly practices. Unfortunately, Fully supervised (FS) segmentation methods, which are known for their efficacy in these tasks, require extensive pixel-level annotations for training. These annotations must be obtained by manually segmenting a large number of images, and they are extremely costly to produce.

\begin{figure}[!t]
  \centering
  \includegraphics[width = 0.9 \textwidth]{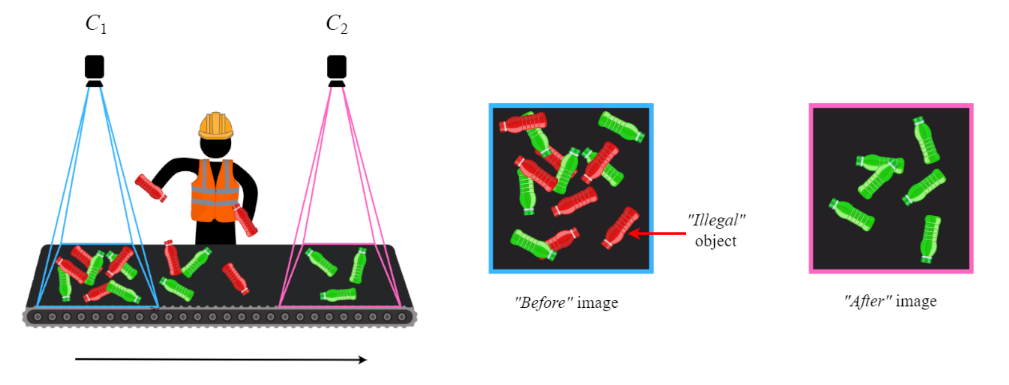}
  \caption{ Two cameras, $C_1$ and $C_2$, are placed along a conveyor belt where a human operator manually removes illegal objects. Camera $C_1$ captures the belt section before the operator’s intervention, while Camera $C_2$ captures the section after, where only legal objects remain. Given a ``before'' image, our goal is to accurately segment objects into two categories: legal objects and illegal objects that should be removed.}
  \label{fig:waste_sorting_scenario}
\end{figure}

In response to these challenges, we present a weakly supervised (WS) solution for waste sorting scenarios like the one illustrated in Fig. \ref{fig:waste_sorting_scenario}, that employs a dual-camera setup along a conveyor belt to streamline the recycling process. One camera captures images of the belt before any manual sorting, while the other captures images after unwanted items have been removed. The goal is to develop a method that automatically segments the objects in these images into two categories: the legal objects that should remain and the illegal ones that need to be removed. The idea proposed in Zerowaste \cite{bashkirova2022zerowaste}, which presents a similar scenario, consists of training an auxiliary classifier to distinguish between ``\textit{before}'' and ``\textit{after}'' images.  Since the ``\textit{before}'' images are characterized by illegal objects that are not present in the ``\textit{after}'' ones, the classifier learns to identify the illegal objects as a distinguishing element for ``\textit{before}'' images. Once this binary classifier has been trained, saliency maps (CAMs)\cite{zhou2016learning} can be used to locate illegal objects in ``\textit{before}'' images: in this way, it is possible to obtain segmentation masks of illegal objects without the need for pixel-wise annotations.
This is a general approach that can leverage any saliency map. Specifically, for the solution exposed in \cite{bashkirova2022zerowaste}, authors use PuzzleCAM \cite{jo2021puzzle}, which generates spatially consistent maps by dividing the image into smaller patches and ensuring consistent activation across these patches. This self-supervised segmentation approach used in \cite{bashkirova2022zerowaste} suffers from two main drawbacks: first, the generated classifier is biased towards the background of the images; secondly, the temporal correlation between saliency maps extracted from consecutive frames of the same camera is not taken into account.

We propose a WS solution that improves the one in \cite{bashkirova2022zerowaste} by exploiting both temporal and spatial coherence. 
Given a collection of videos acquired from both the \textit{before} and \textit{after} cameras, our novel deep-learning framework leverages both temporal and spatial coherence to generate accurate segmentation masks directly from the saliency maps.
More specifically, while training the auxiliary classifier that provides saliency maps of illegal objects, we promote that the saliency maps of the same objects moving in different frames are similar, incorporating in the PuzzleCAM \cite{jo2021puzzle} process a novel reconstruction loss between the map of a central frame $X_t$ and the aggregation of the motion-compensated maps of adjacent frames $X_{t-1}$ and $X_{t+1}$.
Specifically, to adjust the adjacent maps, we employ an optical flow algorithm \cite{wang2020displacement}, which computes the motion between consecutive frames in a video sequence, returning for each pixel both the direction and magnitude of movement.
Our reconstruction loss forces the network to produce identical outputs for seemingly different but conceptually identical frames, allowing our method to accurately highlight and locate illegal objects over time. Consequently, our classifier is trained simultaneously to classify frames in before and after classes, and to provide accurate segmentation masks, ensuring that the network learns to recognize and segment objects based on their temporal dynamics as well as their appearance.

We are the first to leverage a reconstruction loss between saliency maps of nearby frames. To the best of our knowledge, no existing work utilizes this principle at a temporal level. PuzzleCAM \cite{jo2021puzzle} successfully employs this principle from a spatial perspective within a static image but not across multiple frames. Furthermore, to ensure the classifier focuses on the features of the objects rather than on the background, we separate the background from the images, formulating an auxiliary three-class classification problem, rather than the traditional binary classification problem considered in the literature. This results in segmentation masks comprising ``\textit{before}'', ``\textit{after}'', and ``\textit{background}'' pixels. 

Our experiments demonstrate that our approach provides segmentation masks that are very accurate and consistent over time, suitable for industrial waste sorting applications, distinguishing between legal and illegal objects without using any detailed pixel-level annotation.
The paper is organized as follows. Section \ref{sec:related} reviews previous work, Section~\ref{sec:prob_form} formally defines the problem we address and Section~\ref{sec:prop_sol} introduces our approach. Experiments are reported in Section~\ref{sec:experiments} while Section~\ref{sec:conclusions} draws the conclusions.

\section{Related Works}
\label{sec:related} 

The landscape  of segmentation methods based on neural networks can be organized into two main categories: \emph{i})  FS approaches, that require pixel-level annotated datasets for precise segmentation, and \emph{ii}) WS ones, that exploit image-level annotations and are definitely more practical for large-scale applications.
In waste sorting, existing datasets also reflect this 
classification. After presenting an overview of these two categories, we will focus on WS methods that, as in our scenario, take as input videos instead of images.

\paragraph{Fully supervised image segmentation} approaches span a wide range of methods, from simpler region-proposal \cite{girshick2014rich, hariharan2014simultaneous, dai2015convolutional, caesar2016region, shen2020ranet} and fully convolutional networks \cite{noh2015learning, chen2018encoder, adeyinka2019deep, zhang2018exfuse} to transformers \cite{xie2021segformer}. The success of these methods heavily depends on the availability of precise pixel-level annotation for training. In waste sorting, this results in requiring a large training set \( \mathbb{D} = \{(X_i, M_i)\}_{i=1}^N \) composed by RGB images $X_i$, each coupled with its segmentation mask $M_{X_i}$ for different waste classes $\Lambda$, such as cardboard, metal, glass, and plastic, to name a few samples \cite{bashkirova2022zerowaste, 9395690, hong2020trashcan, sanchez2022cleansea, wang2020multi, proencca2020taco}. 

While very useful for general waste sorting, these datasets have significant limitations in real-world applications since very rarely the segmentation task to be addressed in practical applications corresponds to those represented in these datasets. Therefore, one has to reset to manually label several images to train FS methods, which is expensive, time-consuming and challenging, making FS unfeasible especially in facilities that recycle specific materials.

\paragraph{Weakly Supervised image segmentation} techniques remove the need of pixel-level annotations $M$ on images by exploiting various forms of incomplete or imprecise supervision, such as bounding boxes\cite{dai2015boxsup, papandreou2015weakly, khoreva2017simple, song2019box}, scribbles\cite{lin2016scribblesup, vernaza2017learning, tang2018normalized}, and image-level class labels\cite{kolesnikov2016seed, jin2017webly, hou2018self, wei2018revisiting, huang2018weakly, lee2019ficklenet, shimoda2019self, sun2020mining}. 
The latter include Class Activation Maps (CAMs) \cite{zhou2016learning}, which highlight regions of an image that are the most relevant to the class prediction made by a classifier. In this way it is possible to obtain coarse segmentation maps of a specific object using only a classifier trained at image-level. These segmentation masks can be considered as noisy pseudo-labels for training segmentation models\cite{sun2020mining, shimoda2019self, kolesnikov2016seed,huang2018weakly} to get more accurate segmentations.

In recent years, several extensions of CAMs have been developed. Grad-CAM \cite{selvaraju2017grad}, which utilizes the gradients of the target class flowing into the final convolutional layer to generate class-specific saliency maps, is perhaps the most popular solution. Another notable extension, and the one that most inspired this paper, is PuzzleCAM \cite{jo2021puzzle}, which enforces spatial coherence among saliency maps of different patches that constitute the whole image.

In waste sorting, even if generally used for classification tasks, image-based waste datasets \( \mathbb{D} = \{(X, y)_i\}_{i=1}^N \), such as \texttt{TrashNet}\cite{yang2016classification} and \texttt{TrashBox}\cite{kumsetty2022trashbox}, can be used to train WS segmentation networks due to their simple preparation process. However, the waste categories \( y_i \in \Lambda \) (such as glass, paper, cardboard, plastic, metal, and general waste) in these datasets are too broad for industrial needs, which require more detailed distinctions between colored or transparent PET. Furthermore, most datasets are focused on waste images from domestic environments and are thus unsuitable for our industrial setting. 

A notable exception is the ZeroWaste project \cite{bashkirova2022zerowaste}, which, like our work, is collected in a recycling facility. The ZeroWaste dataset is divided into two parts: a widely used supervised component (ZeroWaste-f) and a largely unexplored unsupervised component (ZeroWaste-w). Similarly to our approach, the latter includes images collected “\textit{before}” or “\textit{after}” manual removal from a conveyor belt. Thus it is possible to utilize the WS solution described before. Although the ZeroWaste-w dataset has proposed the method of using saliency maps to identify illegal objects in the ``before'' images, this approach is only sketched and not fully developed in the literature. While we are inspired by this approach, our method extends and refines it. It is widely known that when we use saliency maps for image patches of the same class, the model focuses on key features and only identifies small discriminative parts of a target object \cite{huang2018weakly, wei2017object, zhang2018adversarial}.
To face this challenge and improve the performance, instead of considering static images, we take into account videos and enforce both spatial and temporal coherence. Therefore, our method belongs to the category of WSVS methods described below. 

Another limitation of the ZeroWaste maps, which we address in our solution, is that images of the same class are collected under the same lighting conditions, resulting in a bias of the auxiliary classifier to recognize the class of an image based on the background characteristics rather than on the type of objects present in the image. Such a bias is automatically reflected in the segmentations obtained with saliency maps. In order to overcome this drawback, we take into account saliency maps both on the ``before'' and ``after'' categories, separating the background from the foreground and using it as a third class. It is also worth mentioning that ZeroWaste-w only leverages video data for the ``before'' class, while the ``after'' class consists of static images, and our method requires temporal coherence across both classes to be effective.
\paragraph{Weakly supervised video segmentation} methods consider a whole video sequence \( V = \{X_t\}_{t=1}^T \) annotated with a video label $y$, providing very easy-to-obtain annotation for the neural network. The main difference in using videos $V$ rather than single images $X$ consists in exploiting the rich temporal information available in videos. This 
allows for the propagation of information across frames, which can be exploited to enhance segmentation accuracy and coherence.

Since it is widely known that saliency maps focus on different zones of a single object, activating maps in different frames might highlight different parts of the same objects, due to the different displacement and lighting conditions. For this reason, temporal coherence has been exploited by several approaches to enhance segmentation or localization performance in videos. 

Typically, a classifier is trained on static images to generate saliency maps for individual video frames. These maps are then combined to create comprehensive saliency maps used as supervision for a FS network. Frame-to-Frame (F2F)~\cite{lee2019frame} uses optical flow to warp neighboring maps to a single frame, aggregating them in a post-processing phase to generate detailed maps for the FS network. T-CAM~\cite{belharbi2023tcam} employs a similar process but reuses the auxiliary classifier as an encoder for the segmentation network, overlapping neighboring maps without translation or optical flow compensation. CoLo-CAM~\cite{belharbi2023colo} improves T-CAM by applying a color-based CRF filter on adjacent frames to ensure similar activations in regions with similar colors. In any case, both T-CAM and CoLo-CAM address the task of localization, which is not optimal for our scenario, given the strongly occluded nature of the images we are analyzing. In fact bounding boxes instead of segmentation masks would result in a lot of overlapping, leading to results that would be confused and of limited use. Furthermore, neither T-CAM nor CoLo-CAM uses optical flow. To the best of our knowledge, no existing architecture leverages the advantages of using temporal information during the classifier's training phase. 

In contrast, we combine the principles of F2F\cite{lee2019frame} and PuzzleCAM\cite{jo2021puzzle} and train a classifier directly on videos, forcing it to generate precise and temporally consistent saliency maps by integrating spatial coherence from PuzzleCAM with temporal coherence from video data.  This approach ensures that segmentation masks are as accurate as possible, with the temporal dimension incorporated from the initial training.

\section{Problem formulation}
\label{sec:prob_form}

\begin{figure*}[tb]
  \centering
  \subfloat[]{%
    \includegraphics[width=0.5\textwidth]{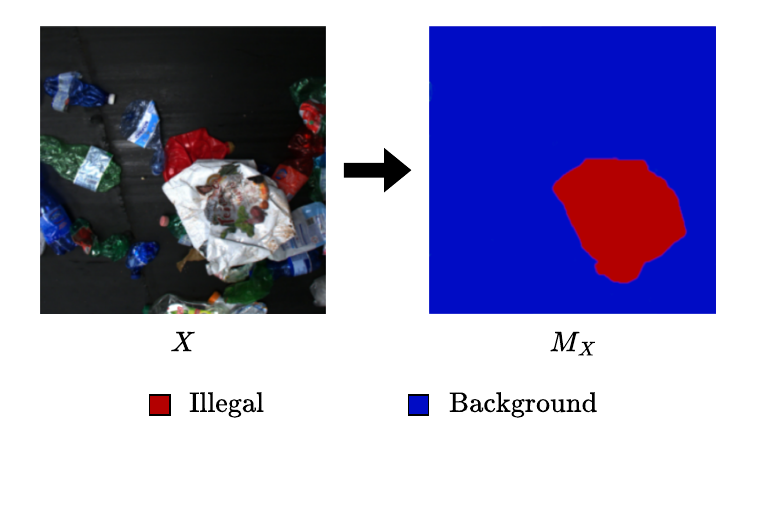}
    \label{fig:illustrated_prob_form-b}}
  \hfill
  \subfloat[]{%
    \includegraphics[width=0.47\textwidth]{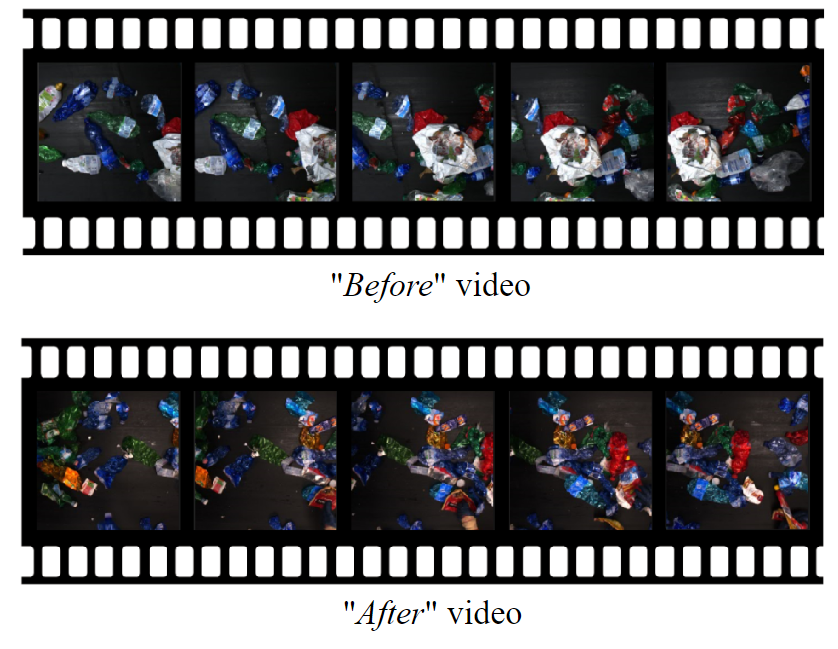}
    \label{fig:illustrated_prob_form-a}}
  \caption{ Problem formulation: (a) an RGB input image \(X\) is processed to generate an accurate output mask \(M_X\). This mask classifies each pixel as illegal (red), or background (blue).
  (b) Training set comprising ``before'' and ``after'' videos. ``Before'' videos capture the conveyor belt before human intervention. ``After'' videos capture the belt after non-colored PET objects have been removed.}
  \label{fig:illustrated_prob_form}
\end{figure*}

We frame our waste sorting problem as a  WS segmentation task where the training data is a set of images collected by the cameras $C_1$ and $C_2$ as shown in Figure \ref{fig:waste_sorting_scenario}. We refer to images collected before the human intervention as ``\textit{before}'' images and to those collected after it as ``\textit{after}'' images. Also, we refer to objects that the human operator must remove as ``\textit{illegal}'' objects while to all the objects that must remain on the belt as ``\textit{legal}'' objects. Given an  RGB image  $ X \in \mathbb{R}^{w \times h \times 3} $ of the conveyor belt, with values normalized between $ [0, 1] $, we aim at segmenting the \textit{illegal} objects that the operator must remove. As illustrated in  Figure~\ref{fig:illustrated_prob_form-b}, this consist in estimating for the image \(X\) a \textit{semantic segmentation mask} \( M_X \in {\Lambda}^{w\times h} \) defined as:
\begin{align}
   M_X(r,c) = y \text{ if pixel} &\text{ at position } (r,c) \text{ in } X
  \text{ belongs to an}\notag \\
  &\text{object } \text{of } \text{class } y \in \Lambda,
  \label{eq:M_X}
\end{align}
where \( {\Lambda} = \{0,1\} \) is the set of \textit{illegal} objects and \textit{background} respectively. Note that in this formulation  \textit{legal} objects are segmented together with  the \textit{background}.

We make the following assumptions:
the training set is composed by videos captured by cameras $C_1$ and $C_2$, 
which are labelled as ``\textit{before}'' or ``\textit{after}'' respectively.
Thus, we are in  a WS setting, \emph{i.e.}, we only know which frames belong to the \textit{before} category and which ones belong to the \textit{after} category.
Formally, the training set $\mathbb{D}$ is defined as follows: \( \mathbb{D} = \{(V, \hat{y})_i\}_{i=1}^N \) denotes a set of $N$ videos, where \( V = \{X_t\}_{t=1}^T \) is an input video with \( T \) RGB frames \(X_t\) defined as above, and \( \hat{y} \in {\Lambda} = \{0, 1\} \) is the class label representing to (\textit{before}, \textit{after}), that is only at video-level (Fig.~\ref{fig:illustrated_prob_form-a}), where (\textit{before}, \textit{after}) videos directly correspond to the presence of (\textit{illegal}, \textit{background}) objects.

\section{Proposed Solution}
\label{sec:prop_sol}

\begin{figure}[tb]
  \centering
  \includegraphics[width = \textwidth]{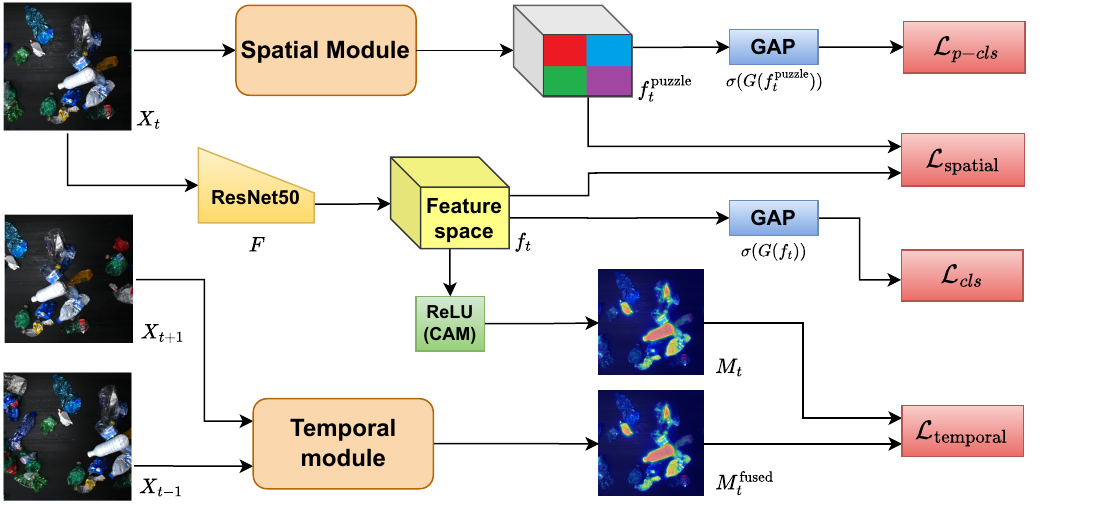}
  \caption{\textbf{Main pipeline illustration}. The overall workflow of our network, which processes a triplet of frames (\(X_{t-1}\), \(X_t\), \(X_{t+1}\)). The spatial module (PuzzleCAM \cite{jo2021puzzle}) outputs a reconstructed feature space $f_t^{puzzle}$ which is pushed to match the original feature space $f_t$ by $\mathcal{L}_{\text{spatial}}$. The temporal module outputs a new saliency map $M_t^{\text{fused}}$ for the central frame $X_t$, obtained from the features of the adjacent frames $X_{t+1}$ and $X_{t-1}$. $M_t^{\text{fused}}$ is then pushed to match the original map $M_t$ by the reconstruction loss $\mathcal{L}_{\text{temporal}}$. $\mathcal{L}_{cls}$ and $\mathcal{L}_{p-cls}$ are instead the classification losses.  The computation of the four losses of the network is described in Sec.~\ref{sec:losses}, while spatial and temporal modules are detailed in Fig.~\ref{fig:spatial_module} and~\ref{fig:temporal_module}, respectively.}
  \label{fig:main_pipeline}
\end{figure}

Inspired by \texttt{zerowaste-w} \cite{bashkirova2022zerowaste}, we train an auxiliary classifier to distinguish between videos taken before and after human intervention.  The classifier learns to identify the \textit{before} video thanks to the presence of illegal objects that are instead absent in the \textit{after} videos.
A saliency map of each individual \textit{before} frame would roughly highlight the regions corresponding to illegal objects, but the resulting segmentation masks might not be very accurate. Thus, to boost the accuracy of saliency maps, we exploit both the spatial and the temporal coherence of the videos, operating on triplets of consecutive frames $X_{t-1}, X_t, X_{t+1}$ as outlined in Fig.~\ref{fig:main_pipeline}. 
As a first step, we remove the background from the images of our dataset in a pre-processing step described in Section~\ref{sec:pre_proc} (Fig. \ref{fig:preprocess}).
As shown in Fig.~\ref{fig:main_pipeline}, the background-removed frames are processed through a pre-trained backbone network (ResNet50) to extract features (Sec.~\ref{sec:feat_extr}), to be handled by two different modules. The \textbf{spatial module} (Sec.~\ref{sec:spatial}) implements the principles of Puzzle-CAM~\cite{jo2021puzzle} and returns the reconstructed feature space \( f_t^{\text{puzzle}} \), obtained by splitting the central frame \( X_t \) into local patches \( X_t^{i,j} \) and by merging back their feature spaces \( f_t^{i,j} \) computed on individual patches (Fig. \ref{fig:spatial_module}).  The \textbf{temporal module} (Sec.~\ref{sec:temporal}) operates along the temporal dimension of videos. It takes as input the adjacent frames \( X_{t-1} \) and \( X_{t+1} \) and, by exploiting optical flow, it reconciles the warped mask $M_{t-1}$ and $M_{t+1}$ and into a central, single, fused $M_{t}^{\text{fused}}$ (Fig. \ref{fig:temporal_module}).
These two modules produce different outputs, which are then compared against the classifier's output with two reconstruction losses. This process forces the classifier to generate consistent saliency maps at spatial and temporal levels (Fig.~\ref{fig:main_pipeline}).

\subsection{Pre-processing}
\label{sec:pre_proc}

All "\textit{before}" images share the same camera \(C_1\), lighting conditions, and belt section, resulting in having all similar backgrounds. The same holds for "\textit{after}" images. However, "\textit{before}" and "\textit{after}" backgrounds are very different from each other. Unfortunately, this condition results in the classifier focusing on the background instead of the features of the objects. For this reason, we preliminary segment foreground objects from the background in our dataset. For both \textit{before} and \textit{after} videos, we estimated a background by computing the pixel-wise median image across all grayscale frames. For each frame, the distance of every pixel with respect to the background estimator is then computed. Pixels significantly different from the estimator are so considered as foreground, resulting in a binary mask that is then applied to the RGB images.

\begin{figure*}[tb]
  \centering
  \subfloat[]{%
    \includegraphics[width=0.49\textwidth]{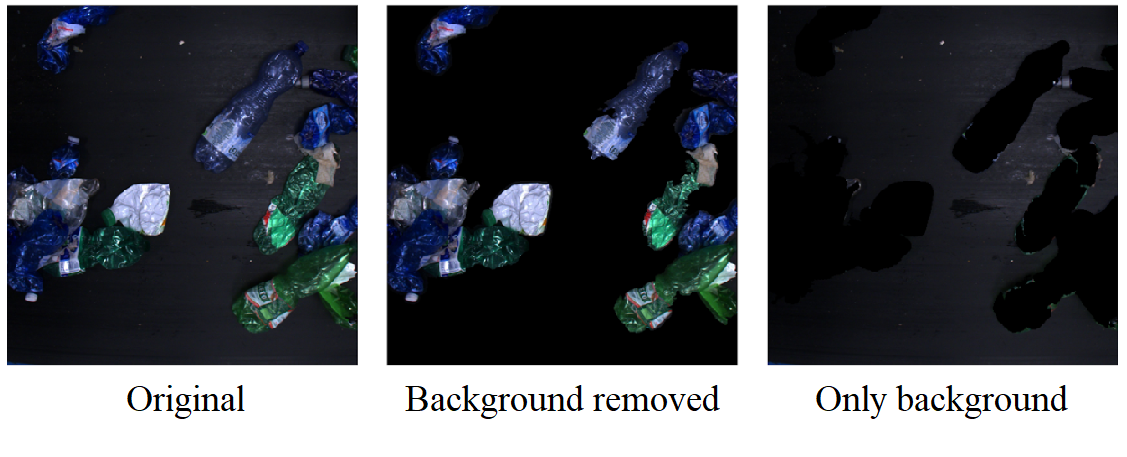}
    \label{fig:bg_only-bg}}
  \hfill
  \subfloat[]{%
    \includegraphics[width=0.49\textwidth]{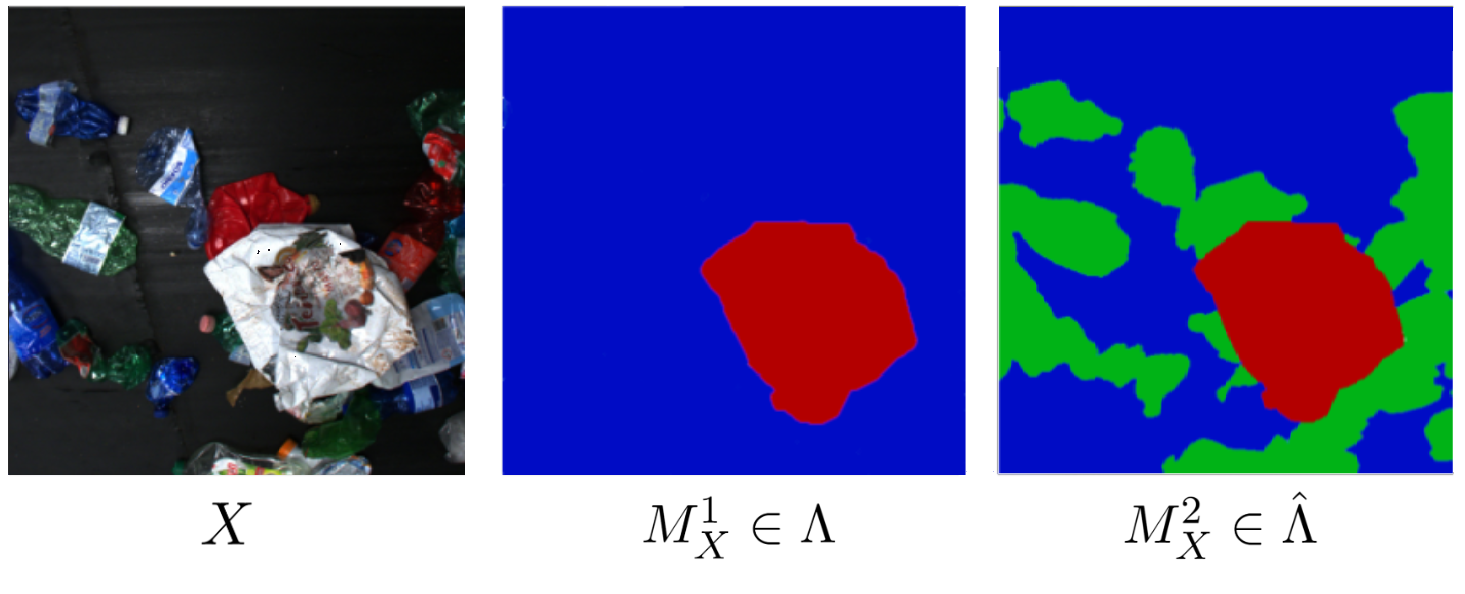}
    \label{fig:domain_shift}}
  \caption{(a) Comparison of images with background, without background and the extracted background itself, which is used to generate a third independent class, respectively.
  (b) By shifting from the $\Lambda$ class domain to the $\hat{\Lambda}$ class domain, we can not only distinguish between \textit{illegal} (red) and \textit{background} (blue) elements, but also segment \textit{legal} (green) objects with a new, more specific, label, distinguishing them from the empty belt regions.}
  \label{fig:preprocess}
\end{figure*}

Then, by inverting the binary masks, we generated a new set of images containing only background elements, expanding the dataset. This results in three classes of images: \textit{after} without background, \textit{before} without background, and only \textit{background} images from both before and after sets. In this way, we drive the classifier to recognize as relevant only the features related to the objects, shifting from the set of classes ${\Lambda}$ = \{0,1\}, corresponding to (\textit{before}, \textit{after}) to $\hat{\Lambda}$ = \{0,1,2\}, corresponding to (\textit{before, after, background}), as shown in Figure \ref{fig:bg_only-bg}.
Figure \ref{fig:domain_shift} illustrates that we use the extracted background itself as a distinct element from legal and illegal objects, which can now be both segmented with specific labels.

\subsection{Feature Extraction and Classification Loss}
\label{sec:feat_extr}
As shown in Fig. \ref{fig:main_pipeline}, each frame \( X_t \) is processed through a pre-trained ResNet50 backbone \( F \) with a classification head $\theta$ that reduces the number of final feature maps to \(|\hat{\Lambda}|\), corresponding to the three classes \textit{after}, \textit{before}, and \textit{background}. The output of the backbone is the feature space \( f_t \):
\begin{equation}
  f_t = F(X_t).
  \label{eq:features}
\end{equation}
which is then processed by a Global Average Pooling (GAP) layer \( G \) to produce the prediction vector \( \hat{z} = \sigma(G(f_t)) \) used for image classification. We utilize a multi-label soft margin loss for this task. For notational convenience, we define \( \bar{z} \) as:
\begin{equation}
  \bar{z} = \begin{cases} 
   \hat{z}, & \text{if } z = 1 \\
    1 - \hat{z}, & \text{otherwise}
    \end{cases}
  \label{eq:z_bar_def}
\end{equation}
and define the classification loss as:
\begin{equation}
  \text{cls}(\hat{z}, z) = -\log(\bar{z})
  \label{eq:class_loss_def}
\end{equation}
where \( z \) is the true label vector of the image \( X_t \) corresponding to its class \( y \in \Lambda \). The classification loss for \( X_t \) is then computed as:
\begin{equation}
  \mathcal{L}_{cls} = \text{cls}(\hat{z}, z)
  \label{eq:L_cls_loss}
\end{equation}
which is used to train the classifier for the image classification task.

\subsection{Spatial Module}
\label{sec:spatial}

\begin{figure}[tb]
  \centering
  \includegraphics[width=\textwidth]{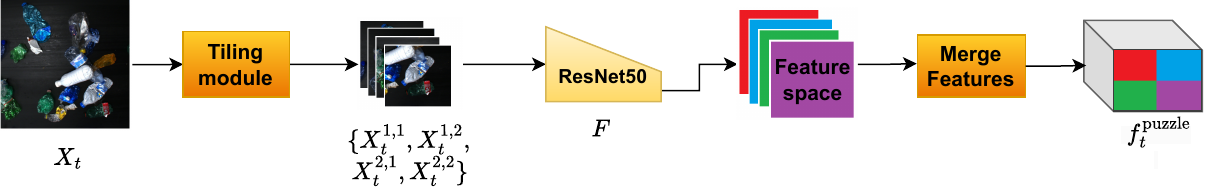}
  \caption{\textbf{Spatial Module}: The central frame \(X_t\) is divided into non-overlapping patches by the tiling module, and for each patch, we extract its feature maps. These sub-feature maps are then re-merged to create a single reconstructed feature space that is compared with the one of the original image $X_t$ through the reconstruction loss $\mathcal{L_{\text{spatial}}}$. This module is the implementation of PuzzleCAM and it aims to improve segmentation by focusing on the spatial arrangement of objects within a single frame.}
  \label{fig:spatial_module}
\end{figure}

Following PuzzleCAM\cite{jo2021puzzle}, our architecture (Fig.~\ref{fig:spatial_module}) is designed to promote spatial coherence of saliency maps when extracting features from a single image as follows.
The Spatial Module processes the central frame \( X_t \) to match its features with those extracted from its patches. More specifically, from an input image \( X_t \) of size \( w \times h \), the tiling module generates non-overlapping tiled patches \( \{X_t^{1,1}, X_t^{1,2}, X_t^{2,1}, X_t^{2,2}\} \) of size \( \frac{w}{2} \times \frac{h}{2} \). Next, we extract \( f_t^{i,j} \) feature spaces for each \( X_t^{i,j} \) as described in~\eqref{eq:features}. Finally, the merging module assembles all \( f_t^{i,j} \) into a single feature space \( f_t^{\text{puzzle}} \) that has the same shape as \( f_t \), the feature space of the original image \( X_t \) (Fig.~\ref{fig:spatial_module}). Using the GAP layer $G$ described in Section \ref{sec:feat_extr}, we map \( f_t^{\text{puzzle}} \) into a prediction vector \( \hat{z}^{\text{puzzle}} = G(f_t^{\text{puzzle}}) \). Using~\eqref{eq:z_bar_def} and \eqref{eq:class_loss_def} we compute a new classification loss as
\begin{equation}
  \mathcal{L}_{p-cls} = \text{cls}(\hat{z}^{\text{puzzle}}, z),
  \label{eq:L_p-cls_loss}
\end{equation}
that improves the image classification performance.
To ensure the classifier produces spatially consistent CAMs, we incorporate a reconstruction loss, which aligns the original and reconstructed feature spaces. This loss is defined as:
\begin{equation}
\mathcal{L}_{\text{spatial}} = \| f_t - f_t^{\text{puzzle}} \|_1.
\label{eq:L_spatial}
\end{equation}

\subsection{Temporal Module}
\label{sec:temporal}

\begin{figure}[tb]
  \centering
  \includegraphics[width = \textwidth]{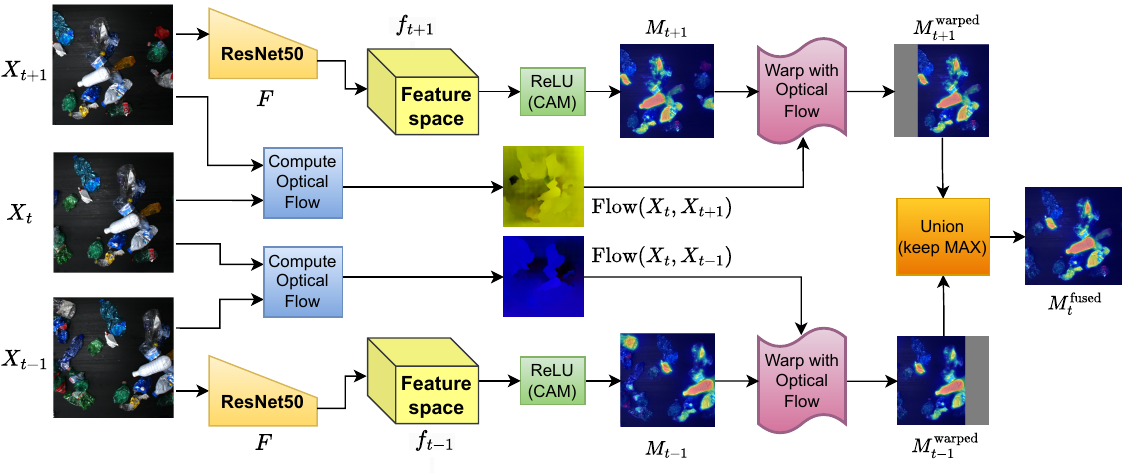}
  \caption{\textbf{Temporal module}: It processes two frames \(X_{t-1}\) and \(X_{t+1}\) adjacent to \(X_t\), to extract their saliency maps. These maps are then warped using optical flow to align them temporally and fused in $M_t^{\text{fused}}$ keeping the pixel-wise maximum values. $M_t^{\text{fused}}$ is then compared with the map of the central \(X_t\) through the reconstruction loss $\mathcal{L}_{\text{temporal}}$. This process ensures that the activations of objects are temporally coherent, promoting the network to segment objects consistently across frames.}
  \label{fig:temporal_module}
\end{figure}

The main contribution introduced by our work is the temporal module, through which we compute temporal consistent saliency maps: this module processes a triplet of frames \( X_{t-1} \), \( X_t \), and \( X_{t+1} \) in a joint classification network employing temporal coherence between the saliency maps of the frames (Fig.~\ref{fig:temporal_module}).

\paragraph{CAM generation.} First, from \( X_{t-1} \) and \( X_{t+1} \), we extract feature spaces \(f_{t-1}\) and \(f_{t+1}\) as in ~\eqref{eq:features}.
Then, for every frame of the triplet \(X_{t-1}, X_t, X_{t+1}\), as done in \cite{jo2021puzzle}, we use a ReLU activation function to compute the saliency map $M$ for the class $y$ the input images belong to:
\begin{equation}
  M = \text{ReLU}(f[y]),
  \label{eq:CAM_def}
\end{equation}
where $f[y]$ represent the $y$-th channel of a feature space $f$.
The computed map $M$ is then normalized by dividing it by its maximum value.
We produce the saliency maps for every frame in triplet \( X_{t-1} \), \( X_t \), and \( X_{t+1} \) obtaining \( M_{t-1} \), \( M_t \), and \( M_{t+1} \), respectively.

\paragraph{Optical Flow Warping and CAM Fusion.} As next step, we align the saliency maps using DICL-FLow \cite{wang2020displacement}, and in practice we compute the optical flows between consecutive frames $X_t$ and $X_{t \pm 1}$ and use these flows to warp the lateral maps to the central one:
\begin{equation}
  \begin{aligned}
    M_i^{\text{warped}} &= \text{Warp}(M_i, \text{ Flow}(X_t, X_i)) \\
    &\text{for } i \in \{t-1, \text{ }t+1\},
  \end{aligned}
  \label{eq:Cam_warped}
\end{equation}
where \( \text{Flow}(X_t, X_i) \) denotes the optical flow between frame \( X_t \) and frame \( X_i \). 
The warped maps for frames \( X_{t-1} \) and \( X_{t+1} \)  are then fused keeping the pixel-wise maximum:
\begin{equation}
  M_t^{\text{fused}} = \max(M_{t-1}^{{\text{warped}}},M_{t+1}^{{\text{warped}}})
  \label{eq:CAM_fused}
\end{equation}
In this way, we obtain a new saliency map for the central frame  \( X_t \) based on the activation of the features identified in the lateral frames.
Figure \ref{fig:temporal_module} illustrates the computation of optical flow, warping of the maps and union of them in blue, purple and dark orange modules, respectively.

\paragraph{Temporal Module Losses.} Using the same principle as PuzzleCAM, we add a reconstruction loss to force the original saliency map \( M_t \) of central frame \(X_t\) to be closed of the reconstructed one from the adjacent frames \(M_t^{\text{fused}}\), namely: 
\begin{equation}
  \mathcal{L}_{\text{temporal}} = \| M_t - M_t^{\text{fused}} \|_1.
  \label{eq:Loss_re-fused}
\end{equation}
In this way, the activations of an object in different positions are temporally coherent.

\subsection{Final Loss Design}
\label{sec:losses}
To summarize, as illustrated in Fig.~\ref{fig:main_pipeline}, we train our network by minimizing a loss function that combines the losses from both the Spatial (PuzzleCAM) and the Temporal modules.
The final loss function \( \mathcal{L}_{\text{total}} \) is the sum of the classification losses given by Eq.~\eqref{eq:L_cls_loss} and \eqref{eq:L_p-cls_loss}, and the reconstruction losses given by~\eqref{eq:L_spatial} and~\eqref{eq:Loss_re-fused}, namely:
\begin{equation}
  \mathcal{L}_{\text{total}} =  \mathcal{L}_{cls} + \mathcal{L}_{p-cls} + \alpha \mathcal{L}_{\text{spatial}} + \beta \mathcal{L}_{\text{temporal}}.
  \label{eq:total_loss_specific}
\end{equation}
where \(\alpha\) and  \(\beta\) are regularization terms that weight respectively the spatial and temporal coherence components given.

\section{Experiments}
\label{sec:experiments}
This section is devoted at assessing the benefits of our solution on both segmentation and classification tasks. After describing our dataset, we present a comparative analysis of our approach against baseline methods, demonstrating the effectiveness of exploiting both temporal coherence in segmentation and background removal in classification tasks.

\subsection{Datasets and Competitors}

We evaluate our method on our custom-collected dataset (named SERUSO and available upon request), which consists of 3682 images, divided into 36 videos for the ``after'' class and 32 videos for the ``before'' class. Specifically, there are 1836 ``after'' images and 1846 ``before'' images, each with a resolution of $2400 \times 2400$ pixels. Cameras have been installed to monitor a conveyor belt containing objects made from PET materials, including transparent, bluish and opaque PET.  The operators remove any object but semi-transparent colored PET ones. As a result, the ``before'' images captured the initial, mixed material flow, while the ``after'' images contain primarily semi-transparent colored PET objects with occasional anomalies (see Fig.~\ref{fig:qualitative} for an example). A total of 364 images were manually labeled by segmenting ``illegal'' objects in the ``before'' images. These segmentation masks were used exclusively for testing. We also performed additional experiments on the Zerowaste-w dataset \cite{bashkirova2022zerowaste}, extending the range of analyzed methods beyond those used by the authors. We benchmarked our method against the approach used by the authors of Zerowaste-w, namely PuzzleCAM \cite{jo2021puzzle}, as well as other CAM-based methods, including Grad-CAM and its extension incorporating temporal coherence (Frame-to-frame \cite{lee2019frame}). 
In addition, we conducted ablation studies to assess the impact of each component of our method.

\begin{table}[tb]
    \caption{Comparison of mIoU scores for different methods across the ZEROWASTE and SERUSO datasets. The superior performance of Frame-2-Frame over GradCAM demonstrates how the benefit is evident even using temporal coherence only in the post-processing phase. On the SERUSO dataset, our model, which incorporates both temporal and spatial coherence during training, outperforms all other CAM-based methods, including PuzzleCAM.
    }
    \centering
    \begin{tabular}{|c|c|c|c|c|}
        \hline
        \textbf{IoU} & \textbf{GradCAM} & \textbf{Frame-2-Frame} & \textbf{PuzzleCAM} & \textbf{Our} \\ \hline
        \textbf{SERUSO} & 22.08 & 27.94 & 34.20 & 37.84 \\ \hline
        \textbf{ZEROWASTE} & 23.13 & 26.43 & 29.87 & Impossible \\ \hline
    \end{tabular}

    \label{tab:comparison}
\end{table}

\begin{table}[tb]
    \caption{IoU scores of models trained with different reconstruction loss configurations—none, only temporal, only spatial, and both—show significant performance improvements with spatial and temporal coherence. On the SERUSO dataset, the temporal module alone greatly enhances performance compared to no module but falls short of the spatial-only module (PuzzleCAM). While the spatial module outperforms the temporal module, combining both yields the highest performance, demonstrating their complementary benefits.}
    \centering
    \begin{tabular}{|c|c|c|c|c|}
        \hline
        \textbf{IoU} & \textbf{None} & \textbf{Only Temporal} & \textbf{Only Spatial} & \textbf{Both} \\ \hline
        \textbf{SERUSO} & 24.08 & 29.23 & 34.20 & 37.84 \\ \hline
    \end{tabular}
    \label{tab:comparison_temp_spat}
\end{table}

\subsection{Results}
All experiments were conducted on a workstation equipped with an Nvidia RTX A6000 GPU. The images were re-scaled to 512 × 512 as the network inputs and the dataset was split into training and validation sets with an 80\% and 20\% split, respectively. In all experiments, \(\alpha\) and \(\beta\) are set to 0 for the first epoch and then linearly increased to a maximum of 4 by the midpoint of training, gradually prioritizing reconstruction losses over classification losses.

\begin{figure*}[tb]
  \centering
  \subfloat[GradCAM]{%
    \includegraphics[width=0.24\textwidth]{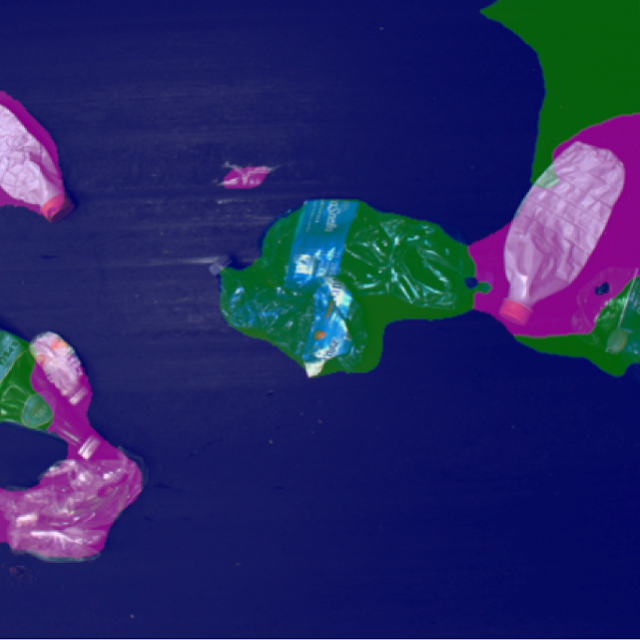}
    \label{fig:qualitative_grad}}
  \hfill
  \subfloat[PuzzleCAM]{%
    \includegraphics[width=0.24\textwidth]{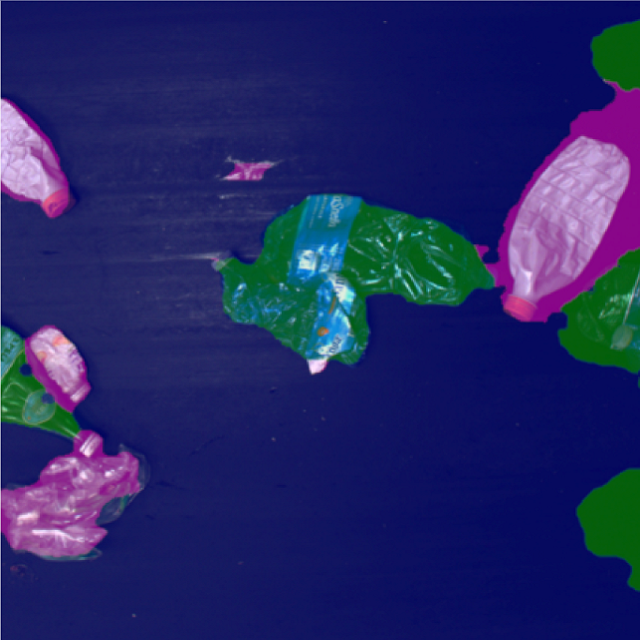}
    \label{fig:qualitative_puzzle}}
  \subfloat[Ours]{%
    \includegraphics[width=0.24\textwidth]{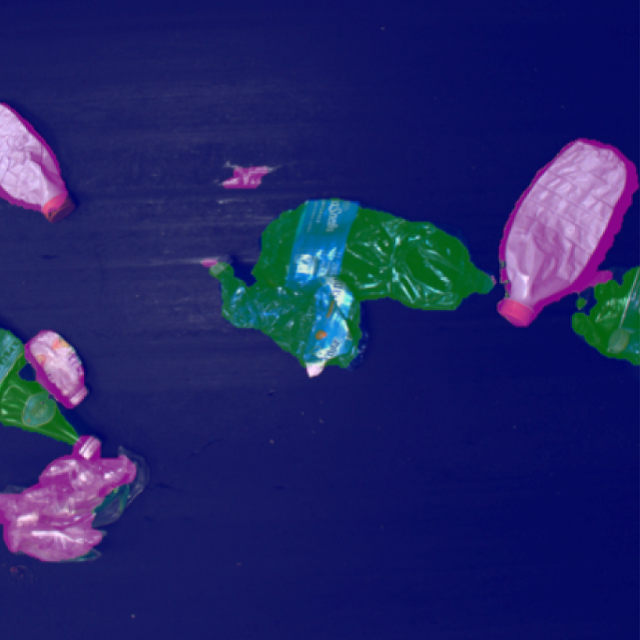}
    \label{fig:qualitative_pof}}
  \subfloat[Ground Truth]{%
    \includegraphics[width=0.24\textwidth]{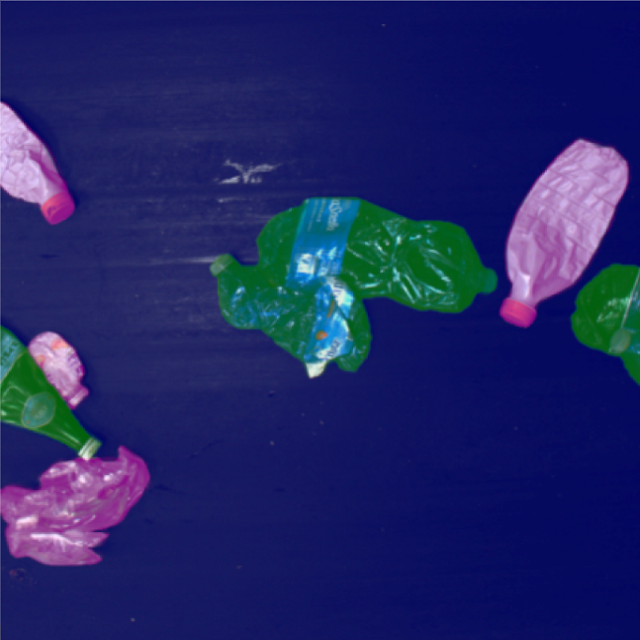}
    \label{fig:qualitative_gt}}
  \caption{Qualitative comparison of segmentation results on the SERUSO dataset, showing how our method (c) attains more precise segmentation compared to both GradCAM (a) and PuzzleCAM (b) thanks to the combined exploitation of spatial and temporal coherence.}
  \label{fig:qualitative}
\end{figure*}

\paragraph{Segmentation.}
To assess the segmentation performance of our saliency maps, we computed the mean Intersection-over-Union (mIoU) over the ``before'' class. Saliency map-based methods identify the most relevant regions for a classifier to assign an image to a specific class. Therefore, all experiments focused on segmenting \textit{illegal} objects in ``\textit{before}'' images.
While confirming that advanced methods like PuzzleCAM show substantial improvements over traditional techniques like GradCAM, Tab.~\ref{tab:comparison} also demonstrates how techniques utilizing temporal coherence significantly enhance segmentation performance. In particular, our method outperforms all others on the SERUSO dataset, showcasing the superiority of integrating temporal coherence directly in the training phase for the segmentation task. Figure \ref{fig:qualitative} shows an example of the qualitative results obtained on SERUSO dataset, highlighting the differences in segmentation performance among various methods.

Unfortunately, it is impossible to train our method on Zerowaste since it provides only static data for the ``after'' class, preventing the incorporation of temporal coherence in training. Table~\ref{tab:comparison_temp_spat} shows how our method performs on the various modules when considered individually. This ablation study demonstrates that both the Spatial and Temporal modules alone outperform a simple saliency map from a classifier trained with only classification losses, whereas their combined use surpasses both the individual configurations.

\paragraph{Classification.}
In order to assess the impact of background removal, we evaluate the classification accuracy of a standard classifier (ResNet 50) on two versions of the Zerowaste-w dataset, one with the original images and one with images having the background removed, as explained in Section \ref{sec:pre_proc}.
Results, in Tab.~\ref{tab:classifier_performance}, show that no background-trained classifier performs well on both the scenarios (with and without background), while the classifier trained on the datasets with background performs well on its training set, but badly on the other, demonstrating the bias given by the background.

\begin{table}[t!]
\caption{Classifier accuracy on the Zerowaste-w dataset w and w/o backgrounds. The classifier trained on data with backgrounds performs excellently when tested on a dataset with backgrounds, but its performance drops when tested on a dataset w/o backgrounds. Conversely, the classifier trained on data without backgrounds achieves similar high performance when tested on both datasets with and w/o backgrounds.
}
\centering
\begin{tabularx}{\textwidth}{|c|>{\centering\arraybackslash}X|>{\centering\arraybackslash}X|>{\centering\arraybackslash}X|>{\centering\arraybackslash}X|}
\hline
\textbf{CLASSIFIER $\rightarrow$} & \multicolumn{2}{c|}{\textbf{BACKGROUNDS}} & \multicolumn{2}{c|}{\textbf{NO BACKGROUNDS}}\\ \hline
\textbf{DATASET $\downarrow$} & Train & Val & Train & Val\\ \hline
BACKGROUNDS &  100 & 99.69 & 98.25 & 96.12 \\ \hline
NO BACKGROUNDS &  68.97 & 64.62 & 99.57 & 98.02 \\ \hline
\end{tabularx}
\label{tab:classifier_performance}
\end{table}

\section{Conclusions}
\label{sec:conclusions}
We addressed the challenging task of industrial waste sorting using a WSVS approach. We proposed a novel method that drives a classifier to produce temporal consistent saliency maps for objects appearing in different frames.  Experiments and ablation studies demonstrated that the use of temporal coherence directly in the classifier's training phase effectively improves the classifier's ability to generate saliency maps, outperforming the mIoU of other saliency maps-based methods. The results obtained in our dual-camera setup are very promising and suggest that this approach can be applied to other industrial processes with similar settings, where it is necessary to manually separate specific objects from a heterogeneous stream e.g. in product quality control processes, where anomalous elements need to be removed from a stream of objects, such as damaged or faulty products. Given that saliency maps are currently computed during the inference phase using only a single frame, future work explores including adjacent frames in the map computation at inference time as well.  Also, as a next step, the segmentation masks obtained can be used as pseudo-labels to supervise a FS segmentation network, to improve the segmentation performance.

\medskip
\noindent
{\small
\textbf{Acknowledgements:}
This paper is supported by PNRR-PE-AI FAIR project funded by the NextGeneration EU program and by GEOPRIDE ID: 2022245ZYB, CUP: D53D23008370001 (PRIN 2022 M4.C2.1.1 Investment).}
%
%
\bibliographystyle{splncs04}
\bibliography{main}
\end{document}